\documentclass{article}

\PassOptionsToPackage{numbers, compress}{natbib}
 \usepackage[preprint]{neurips_2026}

\usepackage[utf8]{inputenc} % allow utf-8 input
\usepackage[T1]{fontenc}    % use 8-bit T1 fonts
\usepackage{hyperref}       % hyperlinks
\usepackage{url}            % simple URL typesetting
\usepackage{booktabs}       % professional-quality tables
\usepackage{amsfonts}       % blackboard math symbols
\usepackage{nicefrac}       % compact symbols for 1/2, etc.
\usepackage{microtype}      % microtypography
\usepackage{xcolor}         % colors
\usepackage{amsmath, amssymb}
\usepackage{graphicx}
\usepackage{booktabs}
\usepackage{wrapfig}
\usepackage{caption}
\title{Clinical Trajectory Alignment for Medical Vision-Language Pre-training}

\author{%
\textbf{Huimin Yan}\textsuperscript{1},
\textbf{Xian Yang}\textsuperscript{2},
\textbf{Zhi Wang}\textsuperscript{1},
\textbf{Liang Bai}\textsuperscript{1}
\\[-1pt]
\textsuperscript{1}Institute of Intelligent Information Processing,
Shanxi University, Taiyuan, China
\\
\textsuperscript{2}Alliance Manchester Business School,
The University of Manchester, Manchester, UK
\\[2pt]
\small
\texttt{yanhm0925@163.com},
\texttt{xian.yang@manchester.ac.uk}
\\[-1pt]
\texttt{202322407045@email.sxu.edu.cn},
\texttt{bailiang@sxu.edu.cn}
}

\begin{document}

\maketitle

\begin{abstract}
Medical vision-language pre-training largely follows a visit-level image-report matching paradigm, aligning paired images and reports at individual visits. While effective for static cross-modal correspondence, this paradigm provides limited supervision for longitudinal clinical change, such as whether abnormalities improve, remain stable, or worsen over time. Learning such change is challenging because temporal semantics are implicit in free-text reports, and different abnormalities within the same patient may evolve asynchronously or even in opposite directions. 
We propose MedCTA, which reframes medical vision-language pre-training from visit-level cross-modal matching to learning clinical change. Rather than compressing a patient history into a single temporal representation, MedCTA models clinical change at two complementary scopes. At the abnormality scope, clinically grounded queries construct abnormality-conditioned visual and textual trajectories to capture heterogeneous abnormality evolution. At the patient-course scope, global image and report sequences are modeled to capture overall clinical progression beyond any individual abnormality. Structured trend supervision is extracted from longitudinal reports by an offline LLM parser, removing the need for manual temporal annotations. Combined with static image-report alignment, MedCTA learns representations that preserve visit-level cross-modal correspondence while encoding longitudinal change semantics. Experiments on temporal image classification, image-text retrieval, and zero-shot classification show consistent gains over strong medical vision-language baselines.
\end{abstract}

\section{Introduction}
\label{sec:introduction}

\begin{figure*}[t]
\centering
\includegraphics[width=1\linewidth]{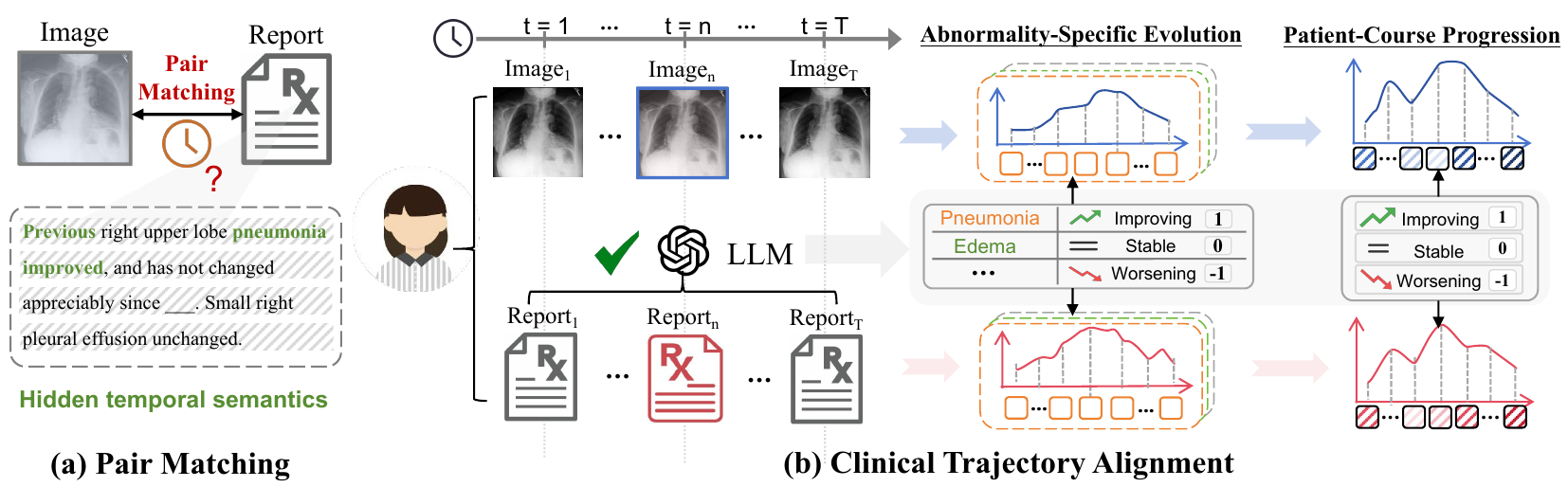}
\caption{
Motivation of MedCTA.
(a) Static image--report matching captures visit-level correspondence but overlooks longitudinal clinical change.
(b) MedCTA exploits follow-up records with LLM-derived trend supervision to model abnormality-specific and patient-course trajectories.
}
\label{fig:mot}
\end{figure*}

Medical vision--language pre-training (VLP) aims to learn clinically meaningful cross-modal representations from medical images and their corresponding reports. 
Existing methods based on image-text contrastive learning~\cite{miao2026biodpp}, masked modeling~\cite{cao2025boosting}, and local region-text matching~\cite{yan2025local} have achieved notable progress in image-text retrieval, disease classification, and phrase grounding~\cite{mei2024medical}.  
However, most existing methods largely follow a visit-level image-report matching paradigm, where paired images and reports are aligned at individual visits~\cite{bluethgen2025vision, AFLoc}. 
Such supervision is effective for learning static cross-modal correspondence, but provides limited guidance for longitudinal clinical change, such as whether abnormalities improve, remain stable, or worsen over time ~\cite{you2025fb, yang2025tempa}. As a result, the temporal evolution semantics contained in patient follow-up records are not explicitly modeled, limiting the ability of current models to learn abnormality progression from longitudinal medical data, as illustrated in Figure~\ref{fig:mot}.

Learning longitudinal clinical change is challenging for two main reasons. 
First, temporal semantics are often implicit in free-text radiology reports rather than available as structured annotations. 
Reports often describe abnormality changes through expressions such as “increased opacity”, “interval improvement”, or “no significant change”, but these descriptions cannot be directly used as supervision without clinical semantic parsing~\cite{cui2025timer, liu2026priorrg}. Second, abnormality progression is inherently heterogeneous. Different abnormalities within the same patient may evolve asynchronously or even in opposite directions. 
Therefore, compressing an entire patient history into a single temporal feature may oversimplify longitudinal abnormality evolution and fail to capture abnormality-specific changes~\cite{zhang2023semi, fan2025medvia}.

To address these challenges, we propose MedCTA, a clinical trajectory alignment framework that reframes medical vision--language pre-training from visit-level cross-modal matching to modeling clinical change. Instead of compressing a patient’s longitudinal history into a single temporal representation, MedCTA explicitly models longitudinal changes at two complementary scopes: abnormality-specific evolution and patient-course progression. At the abnormality scope, MedCTA constructs abnormality-conditioned visual and textual trajectories across visits using clinically grounded abnormality queries. These trajectories capture the evolution of each abnormality over time and are supervised by abnormality-specific trend labels through Abnormality-Specific Trajectory Alignment. At the patient-course scope, MedCTA further models global image and report sequences across visits through Patient-Course Progression Alignment, thereby capturing the patient’s overall clinical progression beyond any individual abnormality. To train these alignment objectives without manual trend annotations, MedCTA adopts an offline large language model to extract structured trend supervision from longitudinal report sequences, including abnormality-specific evolution trends and patient-course progression trends. By jointly optimizing static visit-level image--report alignment, Abnormality-Specific Trajectory Alignment, and Patient-Course Progression Alignment, MedCTA learns cross-modal representations that preserve single-visit image-text correspondence while capturing longitudinal abnormality evolution and patient-course progression.

The main contributions of this paper are summarized as follows:

$\bullet$ We reformulate longitudinal medical VLP as learning from clinical change, extending static visit-level image--report matching to trajectory-level temporal semantic supervision.

$\bullet$ We propose MedCTA, a trajectory alignment framework that models longitudinal change through abnormality-specific trajectories and patient-course progression, thereby capturing heterogeneous abnormality evolution and overall patient-level clinical progression.

$\bullet$ We derive structured trend supervision using an offline LLM parser and demonstrate consistent improvements on temporal classification, image-text retrieval, and zero-shot classification.

\section{Related Work}

\subsection{Medical VLP}
Medical VLP learns transferable cross-modal representations from medical images and radiology reports~\cite{lai2026med,nath2025vila}. Existing methods have made progress through image--text contrastive learning~\cite{MedCLIP,CARZero}, masked modeling~\cite{cheng2023prior}, and fine-grained region--text alignment~\cite{yan2025local, AFLoc}, benefiting tasks such as image--text retrieval, disease classification, phrase grounding, and report understanding~\cite{li2026gmai,chen2025mimo}. However, most methods follow a visit-level image--report matching paradigm, where each pair is treated as an independent static sample. This paradigm learns effective single-visit correspondence but provides limited supervision for longitudinal clinical change, such as improvement, stability, or worsening. Different from prior static medical VLP methods, we reframe pre-training from visit-level cross-modal matching to clinical change learning from longitudinal patient trajectories.

\subsection{Longitudinal Temporal Modeling in Medical VLP}
Longitudinal radiology data contain temporal cues about clinical progression across follow-up examinations~\cite{yang2025tempa, bannur2023learning,liu2025enhanced}. Recent methods have explored such information: ALTA~\cite{lian2025efficient} uses prior X-ray views as temporal context, Med-ST~\cite{DBLP:conf/icml/Yang00W24} introduces cross-modal cycle consistency with historical image--text pairs, and TAMM~\cite{bai2026medical} leverages LLM-derived supervision from adjacent reports. However, these methods mainly capture visual context, temporal correspondence, or short-range local transitions, and thus may overlook heterogeneous abnormality evolution and global patient-course progression. Our method addresses this limitation by modeling clinical change through abnormality-specific trajectories and patient-course progression.

\section{Method}
\label{sec:method}

\subsection{From Static Pair Matching to Change-Aware Pre-training}
\label{subsec:formulation}

Existing medical VLP methods typically align paired images and reports at individual visits. While effective for static cross-modal correspondence, this visit-level matching paradigm provides limited supervision for longitudinal clinical change, such as whether abnormalities improve, remain stable, or worsen over time. MedCTA reframes medical VLP from static pair matching to change-aware pre-training by modeling how clinical states evolve along patient trajectories. 

Formally, we consider a longitudinal medical VLP setting, where each patient is associated with a temporally ordered sequence of image--report pairs:
\begin{equation}
\mathcal{D}=\{\mathcal{S}_i\}_{i=1}^{N}, 
\qquad
\mathcal{S}_i=\{(I_{i,t},R_{i,t},\tau_{i,t})\}_{t=1}^{T_i},
\end{equation}
where $I_{i,t}$ and $R_{i,t}$ denote the medical image and the corresponding radiology report of patient $i$ at the $t$-th visit, $\tau_{i,t}$ is the examination timestamp, and $T_i$ is the number of visits.

MedCTA retains static image--report alignment to preserve visit-level cross-modal correspondence. 
For a batch of $B$ image--report pairs, let $\mathbf{v}_b$ and $\mathbf{r}_b$ denote the global image and report representations of the $b$-th pair, respectively. The static image--report alignment loss is defined as
\begin{equation}
\mathcal{L}_{\mathrm{clip}}
=
\frac{1}{2}
\left(
\mathcal{L}_{I\rightarrow T}
+
\mathcal{L}_{T\rightarrow I}
\right),
\quad
\mathcal{L}_{I\rightarrow T}
=
-\frac{1}{B}
\sum_{b=1}^{B}
\log
\frac{
\exp(\mathrm{sim}(\mathbf{v}_{b},\mathbf{r}_{b})/\tau)
}{
\sum_{j=1}^{B}
\exp(\mathrm{sim}(\mathbf{v}_{b},\mathbf{r}_{j})/\tau)
},
\end{equation}
where $\mathcal{L}_{T\rightarrow I}$ is symmetric, $\mathrm{sim}(\cdot,\cdot)$ is cosine similarity, and $\tau$ is the temperature.

\begin{figure*}[t]
\centering
\includegraphics[width=1\linewidth]{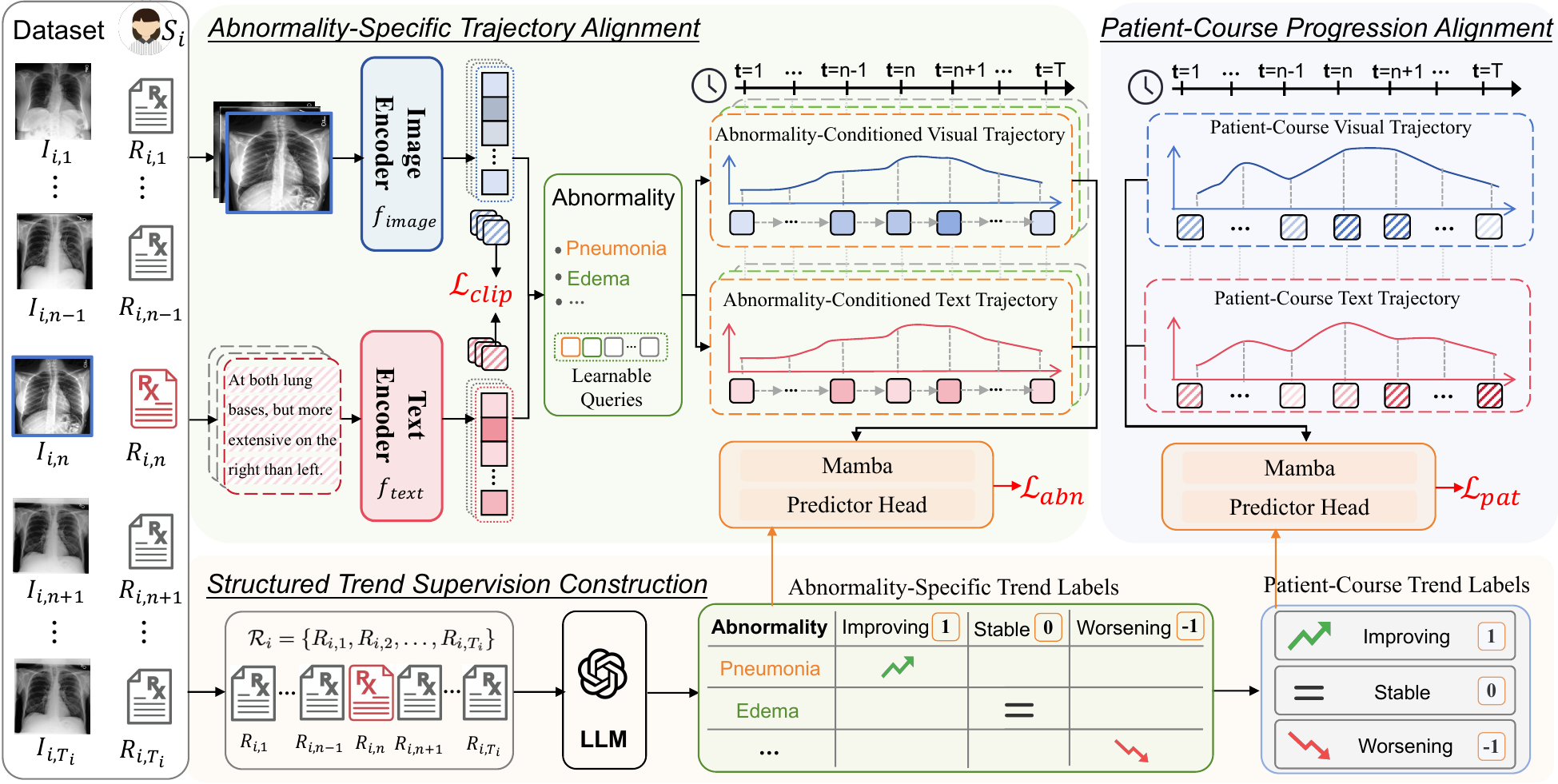}
\caption{
Overview of MedCTA. Given a longitudinal image--report sequence $\mathcal{S}_i$, MedCTA jointly performs static image--report alignment, Abnormality-Specific Trajectory Alignment, and Patient-Course Progression Alignment. 
Abnormality-specific and patient-course trajectories are temporally encoded by Mamba and constrained by LLM-derived trend supervision, enabling semantic-level trajectory alignment for abnormality evolution and overall clinical progression.
}
\label{fig:framework}
\end{figure*}

However, $\mathcal{L}_{\mathrm{clip}}$ only learns visit-level state matching and cannot explicitly supervise longitudinal change. Therefore, MedCTA further introduces two change-aware objectives: Abnormality-Specific Trajectory Alignment for heterogeneous abnormality evolution and Patient-Course Progression Alignment for global clinical progression, as shown in Figure~\ref{fig:framework}.

\subsection{Structured Trend Supervision from Longitudinal Reports}
\label{subsec:trend_supervision}

The temporal semantics of longitudinal radiology reports are usually implicit in free-text reports and cannot be directly used as structured supervision.
To obtain learnable temporal supervision, we use an LLM as an \emph{offline parser} to extract structured trend labels from complete longitudinal report sequences. 
For patient $i$, the longitudinal report sequence is
\begin{equation}
\mathcal{R}_i=\{R_{i,1},R_{i,2},\dots,R_{i,T_i}\}.
\end{equation}
Specifically, MedCTA uses the same offline LLM parser $\mathcal{G}_{\mathrm{LLM}}$ with task-specific prompts to derive two types of trend labels from $\mathcal{R}_i$: abnormality-specific trend labels and patient-course trend labels.
Given a fixed abnormality set $\mathcal{A}=\{a_k\}_{k=1}^{K}$, the abnormality-specific trend label is generated as
\begin{equation}
y_{i,k}^{a}
=
\mathcal{G}_{\mathrm{LLM}}(\mathcal{R}_i,a_k;\pi_{\mathrm{abn}}),
\qquad
y_{i,k}^{a}\in\{-1,0,1\},
\end{equation}
where $\pi_{\mathrm{abn}}$ denotes the abnormality-level prompting instruction. The labels $-1$, $0$, and $1$ indicate worsening, stable condition, and improvement, respectively. 
Since not all abnormalities appear in each patient sequence, we use a validity mask $m_{i,k}\in\{0,1\}$, where $m_{i,k}=1$ if abnormality $a_k$ has an available trend label for patient $i$, and $m_{i,k}=0$ otherwise.

In parallel, the patient-course trend label is generated by the same LLM parser with a patient-level prompt:
\begin{equation}
y_i^{p}
=
\mathcal{G}_{\mathrm{LLM}}(\mathcal{R}_i;\pi_{\mathrm{pat}}),
\qquad
y_i^{p}\in\{-1,0,1\},
\end{equation}
where $\pi_{\mathrm{pat}}$ denotes the patient-course prompting instruction. This label provides sequence-level supervision for patient-course progression modeling.

\subsection{Abnormality-Conditioned Trajectory Construction}
\label{subsec:abnormality_querying}

To model the longitudinal evolution of individual abnormalities, abnormality queries serve as semantic selectors that identify abnormality-relevant visual and textual representations at each visit. These visit-wise representations are then organized across time to form abnormality-conditioned visual and textual trajectories.
Given a predefined abnormality set $\mathcal{A}=\{a_k\}_{k=1}^{K}$, 
we associate each abnormality $a_k$ with a learnable query 
$\mathbf{q}_k\in\mathbb{R}^{d}$, and collect all queries as 
$\mathbf{Q}=[\mathbf{q}_1;\mathbf{q}_2;\cdots;\mathbf{q}_K]\in\mathbb{R}^{K\times d}$.
To inject clinical semantic priors, we initialize each query using the global
$\mathrm{[CLS]}$ embedding of an abnormality-specific prompt $P_k$, e.g.,
"a clinical finding of $a_k$", encoded by the text encoder~\cite{BioClinicalBERT}:
\begin{equation}
\mathbf{q}_k^{(0)}
=
f_{\mathrm{text}}^{\mathrm{CLS}}(P_k),
\qquad
k=1,\dots,K.
\end{equation}

The initialized query $\mathbf{q}_k^{(0)}$ is then treated as the initial value of the learnable parameter $\mathbf{q}_k$ and is updated during pre-training.

To adapt the query to different modalities, we project it into image and text spaces:
\begin{equation}
\mathbf{q}_{k}^{I}
=
\mathbf{W}_{I}\mathbf{q}_k,
\qquad
\mathbf{q}_{k}^{T}
=
\mathbf{W}_{T}\mathbf{q}_k,
\end{equation}
where $\mathbf{W}_{I}$ and $\mathbf{W}_{T}$ are modality-specific projection matrices.

For each visit $(I_{i,t},R_{i,t})$, the image encoder $f_{\text{image}}$~\cite{DosovitskiyB0WZ21} and text encoder $f_{\text{text}}$~\cite{BioClinicalBERT} produce global features for visit-level alignment and local features for abnormality-specific extraction:
\begin{equation}
\label{eq:encoder}
f_{\text{image}}: I_{i,t} \to (\mathbf{g}_{i,t}^{I}, \mathbf{H}_{i,t}^{I}),
\quad
f_{\text{text}}: R_{i,t} \to (\mathbf{g}_{i,t}^{T}, \mathbf{H}_{i,t}^{T}).
\end{equation}
where $\mathbf{g}_{i,t}^{I}$ and $\mathbf{g}_{i,t}^{T}$ denote the global image and text representations, respectively, while $\mathbf{H}_{i,t}^{I}$ and $\mathbf{H}_{i,t}^{T}$ denote the patch-level image features and token-level report features, respectively.

The modality-specific queries attend to local features to obtain abnormality-specific visual and textual states:
\begin{equation}
\mathbf{e}_{i,t,k}^{I}
=
\mathrm{Attn}
(
\mathbf{q}_{k}^{I},
\mathbf{H}_{i,t}^{I},
\mathbf{H}_{i,t}^{I}
),
\qquad
\mathbf{e}_{i,t,k}^{T}
=
\mathrm{Attn}
(
\mathbf{q}_{k}^{T},
\mathbf{H}_{i,t}^{T},
\mathbf{H}_{i,t}^{T}
),
\end{equation}
where $\mathbf{e}_{i,t,k}^{I}$ and $\mathbf{e}_{i,t,k}^{T}$ denote the image-side and text-side abnormality-specific embeddings of $a_k$ at visit $t$, respectively. In this way, the image and text representations are organized into abnormality-conditioned semantic spaces.

Given the visit-wise abnormality-conditioned representations, MedCTA constructs visual and textual trajectories for each abnormality by organizing these representations across visits. For patient $i$ and abnormality $a_k$, the resulting image-side and text-side abnormality trajectories are defined as:

\begin{equation}
\mathcal{E}_{i,k}^{I}
=
\{\mathbf{e}_{i,1,k}^{I},\mathbf{e}_{i,2,k}^{I},\dots,\mathbf{e}_{i,T_i,k}^{I}\},
\qquad
\mathcal{E}_{i,k}^{T}
=
\{\mathbf{e}_{i,1,k}^{T},\mathbf{e}_{i,2,k}^{T},\dots,\mathbf{e}_{i,T_i,k}^{T}\}.
\end{equation}

\subsection{Abnormality-Specific Trajectory Alignment}
\label{subsec:trajectory_alignment}

Given the abnormality-conditioned trajectories, we use modality-specific Mamba-based temporal encoders~\cite{DBLP:journals/corr/abs-2312-00752} to encode the image-side and text-side evolution of each abnormality:
\begin{equation}
\mathbf{z}_{i,k}^{I,a}
=
g_{\text{Mamba}}^{I,a}(\mathcal{E}_{i,k}^{I}),
\qquad
\mathbf{z}_{i,k}^{T,a}
=
g_{\text{Mamba}}^{T,a}(\mathcal{E}_{i,k}^{T}),
\end{equation}
where $\mathbf{z}_{i,k}^{I,a}$ and $\mathbf{z}_{i,k}^{T,a}$ denote the temporally encoded image-side and text-side trajectory representations of abnormality $a_k$ for patient $i$, respectively. Padding masks are applied during temporal encoding to handle variable-length visit sequences across patients.

Instead of directly matching image and text trajectory embeddings, MedCTA aligns them in a shared clinical-change space supervised by the LLM-derived abnormality trend label $y_{i,k}^{a}$. The image-side and text-side trend predictions are computed as
\begin{equation}
\hat{\mathbf{p}}_{i,k}^{I,a}
=
\mathrm{softmax}
(
h_I^{a}(\mathbf{z}_{i,k}^{I,a})
),
\qquad
\hat{\mathbf{p}}_{i,k}^{T,a}
=
\mathrm{softmax}
(
h_T^{a}(\mathbf{z}_{i,k}^{T,a})
),
\end{equation}
where $h_I^{a}(\cdot)$ and $h_T^{a}(\cdot)$ are image-side and text-side abnormality-specific trend classification heads.

The abnormality-specific trajectory alignment loss is defined as
\begin{equation}
\mathcal{L}_{\mathrm{abn}}
=
\frac{1}{2\sum_{i=1}^{N}\sum_{k=1}^{K}m_{i,k}}
\sum_{i=1}^{N}
\sum_{k=1}^{K}
m_{i,k}
\left[
\mathrm{CE}\left(\hat{\mathbf{p}}_{i,k}^{I,a}, y_{i,k}^{a}\right)
+
\mathrm{CE}\left(\hat{\mathbf{p}}_{i,k}^{T,a}, y_{i,k}^{a}\right)
\right],
\end{equation}
where $\mathrm{CE}(\cdot,\cdot)$ denotes the cross-entropy loss, and $m_{i,k}$ is a validity mask indicating whether abnormality $a_k$ has an available trend label for patient $i$. The label $y_{i,k}^{a}$ is mapped to a three-class trend category, corresponding to worsening, stable, and improving conditions.
This objective encourages image and text trajectories of the same abnormality to be consistent at the trend-semantic level.

\subsection{Patient-Course Progression Alignment}

While abnormality-specific trajectories capture fine-grained evolution, they may not fully reflect the overall patient course. MedCTA therefore models patient-course progression from global longitudinal image and text representations.

For patient $i$, we organize the visit-level global representations from Eq.~\eqref{eq:encoder} into image and text sequences:
\begin{equation}
\mathcal{U}_{i}^{I}
=
\{\mathbf{g}_{i,t}^{I}\}_{t=1}^{T_i},
\qquad
\mathcal{U}_{i}^{T}
=
\{\mathbf{g}_{i,t}^{T}\}_{t=1}^{T_i}.
\end{equation}
To capture long-range patient-course progression, we perform Mamba-based temporal modeling~\cite{DBLP:journals/corr/abs-2312-00752} over the global image and text sequences:
\begin{equation}
\mathbf{z}_{i}^{I,p}
=
g_{\mathrm{Mamba}}^{I,p}(\mathcal{U}_{i}^{I}),
\qquad
\mathbf{z}_{i}^{T,p}
=
g_{\mathrm{Mamba}}^{T,p}(\mathcal{U}_{i}^{T}).
\end{equation}
where $\mathbf{z}_{i}^{I,p}$ and $\mathbf{z}_{i}^{T,p}$ denote the image-side and text-side patient-course progression representations.

The learned patient-course representations are supervised by the LLM-derived patient-course trend label $y_i^{p}$. The image-side and text-side predictions are computed as
\begin{equation}
\hat{\mathbf{p}}_{i}^{I,p}
=
\mathrm{softmax}
(
h_I^{p}(\mathbf{z}_{i}^{I,p})
),
\qquad
\hat{\mathbf{p}}_{i}^{T,p}
=
\mathrm{softmax}
(
h_T^{p}(\mathbf{z}_{i}^{T,p})
),
\end{equation}
where $h_I^{p}(\cdot)$ and $h_T^{p}(\cdot)$ are patient-course trend classification heads. The patient-course progression alignment loss is defined as
\begin{equation}
\mathcal{L}_{\mathrm{pat}}
=
\frac{1}{2N}
\sum_{i=1}^{N}
\left[
\mathrm{CE}\left(\hat{\mathbf{p}}_{i}^{I,p}, y_i^{p}\right)
+
\mathrm{CE}\left(\hat{\mathbf{p}}_{i}^{T,p}, y_i^{p}\right)
\right],
\end{equation}
where $y_i^{p}$ is mapped to a three-class label corresponding to overall worsening, stable condition, and improvement. 
This loss encourages the model to capture the overall clinical progression of each patient across longitudinal visits, thereby enabling patient-course progression alignment.

\subsection{Joint Optimization with Static Image--Report Alignment}
\label{subsec:objective}

MedCTA is trained by jointly optimizing static image--report alignment, abnormality-specific trajectory alignment, and patient-course progression alignment:
\begin{equation}
\mathcal{L}
=
\mathcal{L}_{\mathrm{clip}}
+
\mathcal{L}_{\mathrm{abn}}
+
\mathcal{L}_{\mathrm{pat}},
\end{equation}
where $\mathcal{L}_{\mathrm{clip}}$ preserves visit-level cross-modal correspondence, $\mathcal{L}_{\mathrm{abn}}$ supervises abnormality-specific evolution, and $\mathcal{L}_{\mathrm{pat}}$ models overall patient-course progression.

\section{Experiments}

\subsection{Experimental Setting}

We evaluate MedCTA on both longitudinal temporal reasoning and cross-modal understanding tasks. Below, we briefly describe the pre-training data, downstream benchmarks, and baselines. Additional implementation and evaluation details are provided in Appendix~\ref{Appendix_A}.

\textbf{Pre-training Data.}
We pre-train MedCTA on MIMIC-CXR~\cite{johnson2019mimic} and reorganize image--report pairs into patient-level longitudinal sequences ordered by study time. Following prior temporal VLP settings~\cite{DBLP:conf/icml/Yang00W24}, each sequence contains up to four consecutive visits.

\textbf{Downstream Tasks.}
We evaluate MedCTA on temporal reasoning and cross-modal understanding tasks. Temporal reasoning includes temporal image classification, temporal sentence similarity classification, and dynamic phrase grounding on Chest ImaGenome~\cite{wu2021chest} and MS-CXR-T~\cite{bannur2023learning}. Cross-modal understanding includes image--text retrieval on MIMIC-5$\times$200~\cite{MedCLIP} and zero-shot classification on COVIDx~\cite{wang2020covid} and RSNA Pneumonia~\cite{shih2019augmenting}.

\textbf{Baselines.}
We compare MedCTA with representative medical vision--language methods under the same evaluation settings. The baselines include non-temporal VLP methods, i.e., MGCA~\cite{wang2022multi}, MedCLIP~\cite{MedCLIP}, CARZero~\cite{CARZero}, PRIOR~\cite{cheng2023prior}, and MAVL~\cite{phan2024decomposing}, as well as temporal medical VLP methods, i.e., Med-ST~\cite{DBLP:conf/icml/Yang00W24}, ALTA~\cite{lian2025efficient}, and TAMM~\cite{bai2026medical}. When official code is available, we reproduce results using the released implementations; otherwise, we follow the original protocols with the same backbone and training settings for fair comparison.

\subsection{Experimental Results}
To comprehensively evaluate MedCTA, we divide our evaluation into two categories: temporal trend understanding tasks and standard vision-language benchmarks.

\subsubsection{Temporal Trend Understanding Tasks}

\textbf{Temporal Image Classification.}
To evaluate temporal progression modeling from longitudinal chest X-rays, we fine-tune on Chest ImaGenome and evaluate temporal image classification on MS-CXR-T over five abnormalities: Consolidation, Edema, Pleural Effusion, Pneumonia, and Pneumothorax. 
The task predicts one of three progression categories, i.e., \texttt{improving}, \texttt{stable}, and \texttt{worsening}, for each abnormality.
As shown in Table~\ref{table:temporal_image}, MedCTA achieves superior performance across all diseases. This indicates that explicitly modeling longitudinal clinical change at both abnormality and patient-course levels provides more discriminative representations than visit-level image--report matching alone, especially for distinguishing improvement, stability, and worsening.

\begin{table*}[t]
\centering
\caption{Temporal image classification results. Accuracy (\%) is reported across five abnormality categories. Best results are shown in bold, and second-best results are underlined.}
\label{table:temporal_image}
\small
\setlength{\tabcolsep}{4pt}
\renewcommand{\arraystretch}{1}
\begin{tabular*}{\textwidth}{@{\extracolsep{\fill}}lccccc@{}}
\toprule
\textbf{Method} & \textbf{Consolidation} & \textbf{Edema} & \textbf{Pl. Effusion} & \textbf{Pneumonia} & \textbf{Pneumothorax} \\
\midrule
MGCA          & 44.99 $\pm$ 0.47 & 62.71 $\pm$ 0.24 & 57.17 $\pm$ 0.69 & 63.73 $\pm$ 0.89 & 52.16 $\pm$ 1.02 \\
MedCLIP       & 53.40 $\pm$ 0.87 & 42.85 $\pm$ 0.01 & 49.96 $\pm$ 0.28 & 67.10 $\pm$ 0.22 & 55.46 $\pm$ 0.02 \\
PRIOR         & 41.85 $\pm$ 1.51 & 43.97 $\pm$ 1.03 & 59.21 $\pm$ 1.19 & 41.41 $\pm$ 3.24 & 45.33 $\pm$ 2.13 \\
MAVL          & 43.79 $\pm$ 0.00 & 42.85 $\pm$ 0.00 & 48.66 $\pm$ 0.00 & 67.10 $\pm$ 0.00 & 55.46 $\pm$ 0.00 \\
CARZero       & 48.95 $\pm$ 1.25 & 51.92 $\pm$ 1.64 & 50.19 $\pm$ 1.72 & 57.44 $\pm$ 1.35 & 47.84 $\pm$ 1.78 \\
Med-ST        & 60.57 $\pm$ 1.18 & 67.35 $\pm$ 0.32 & 58.47 $\pm$ 1.50 & 65.00 $\pm$ 0.34 & 54.18 $\pm$ 0.81 \\
ALTA          & 52.76 $\pm$ 1.86 & 66.92 $\pm$ 0.89 & 50.92 $\pm$ 1.72 & 66.82 $\pm$ 0.19 & 51.04 $\pm$ 0.41 \\
TAMM & \underline{65.51 $\pm$ 0.83} & \underline{69.22 $\pm$ 0.01} & \underline{63.18 $\pm$ 0.45} & \underline{72.71 $\pm$ 0.77} & \underline{60.50 $\pm$ 0.01} \\
\textbf{Ours} & \textbf{67.14 $\pm$ 0.82} & \textbf{70.30 $\pm$ 0.16} & \textbf{65.59 $\pm$ 1.03} & \textbf{72.85 $\pm$ 0.35} & \textbf{61.66 $\pm$ 0.05} \\
\bottomrule
\end{tabular*}
\end{table*}

\begin{wrapfigure}{r}{0.52\textwidth}
    % \vspace{-0.8\baselineskip} 
    \centering
    \includegraphics[width=\linewidth]{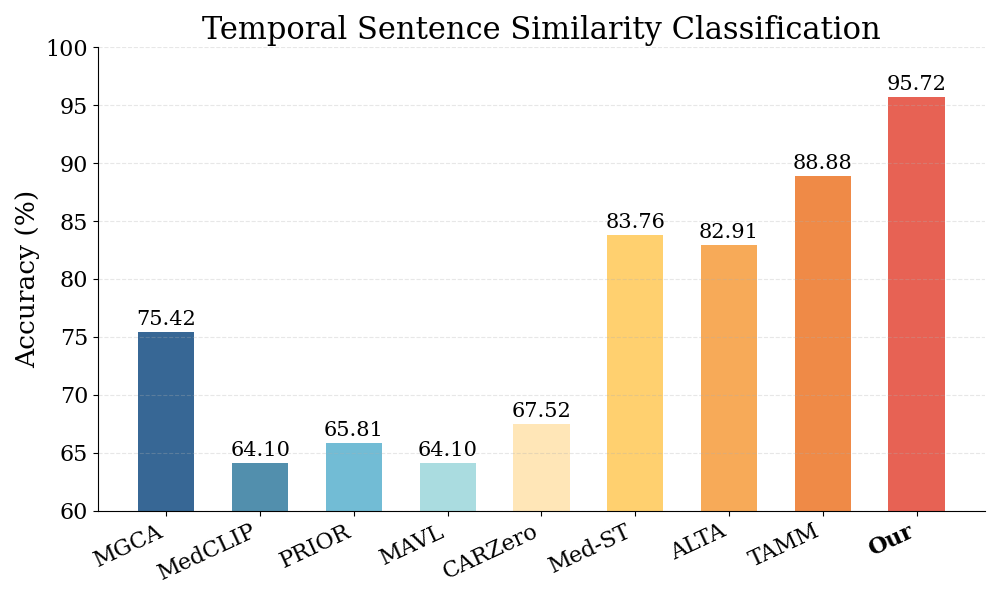}
    \caption{
    Accuracy (\%) for temporal sentence similarity classification.
    }
    \label{fig:temporal_sentence_acc}
    \vspace{-0.5\baselineskip} 
\end{wrapfigure}

\textbf{Temporal Sentence Similarity Classification.}
To evaluate the temporal language understanding ability of MedCTA, we conduct zero-shot sentence similarity classification on the MS-CXR-T dataset. In particular, this experiment examines the text encoder of each model to determine how well it captures semantic shifts and temporal progression in textual descriptions.
Fig.~\ref{fig:temporal_sentence_acc} shows that MedCTA achieves the best zero-shot accuracy of 95.72\%. This result indicates that MedCTA learns more temporally aware textual representations, enabling better recognition of progression-related semantics such as improvement, stability, and worsening in reports.

\begin{figure*}
\centering
\includegraphics[width=0.98\linewidth]{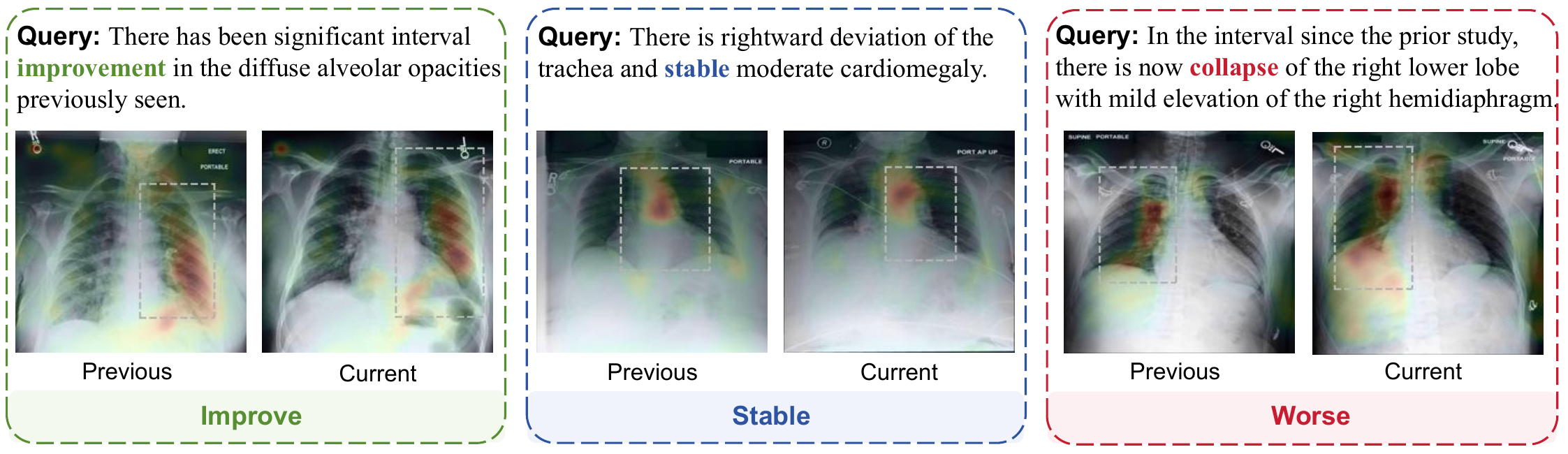}
\caption{
Qualitative dynamic phrase grounding results on Chest ImaGenome for improving, stable, and worsening cases. The left and right images denote the prior and current studies, respectively, with highlighted regions indicating visual evidence related to temporal progression.
}
\label{fig:temporal_dynamic_grounding}
\end{figure*}

\textbf{Dynamic Phrase Grounding.} We evaluate dynamic phrase grounding using Chest ImaGenome bounding-box annotations, aiming to localize temporal progression descriptions across two consecutive chest X-rays. 
Specifically, we compute cosine similarity between patch-level image embeddings and temporal description embeddings, and upsample the patch-wise similarity scores to generate grounding heatmaps. 
Fig.~\ref{fig:temporal_dynamic_grounding} presents representative examples from three progression categories, i.e., \texttt{improving}, \texttt{stable}, and \texttt{worsening}, with the left and right images denoting the prior and current studies, respectively.
The results show that MedCTA can better localize regions associated with longitudinal changes by organizing abnormality-relevant visual representations across visits.

% % % % % % % % % % % % % % % % % % % % % 

\subsubsection{Standard Vision-Language Benchmarks}

\begin{figure*}
\centering
\includegraphics[width=0.9\linewidth]{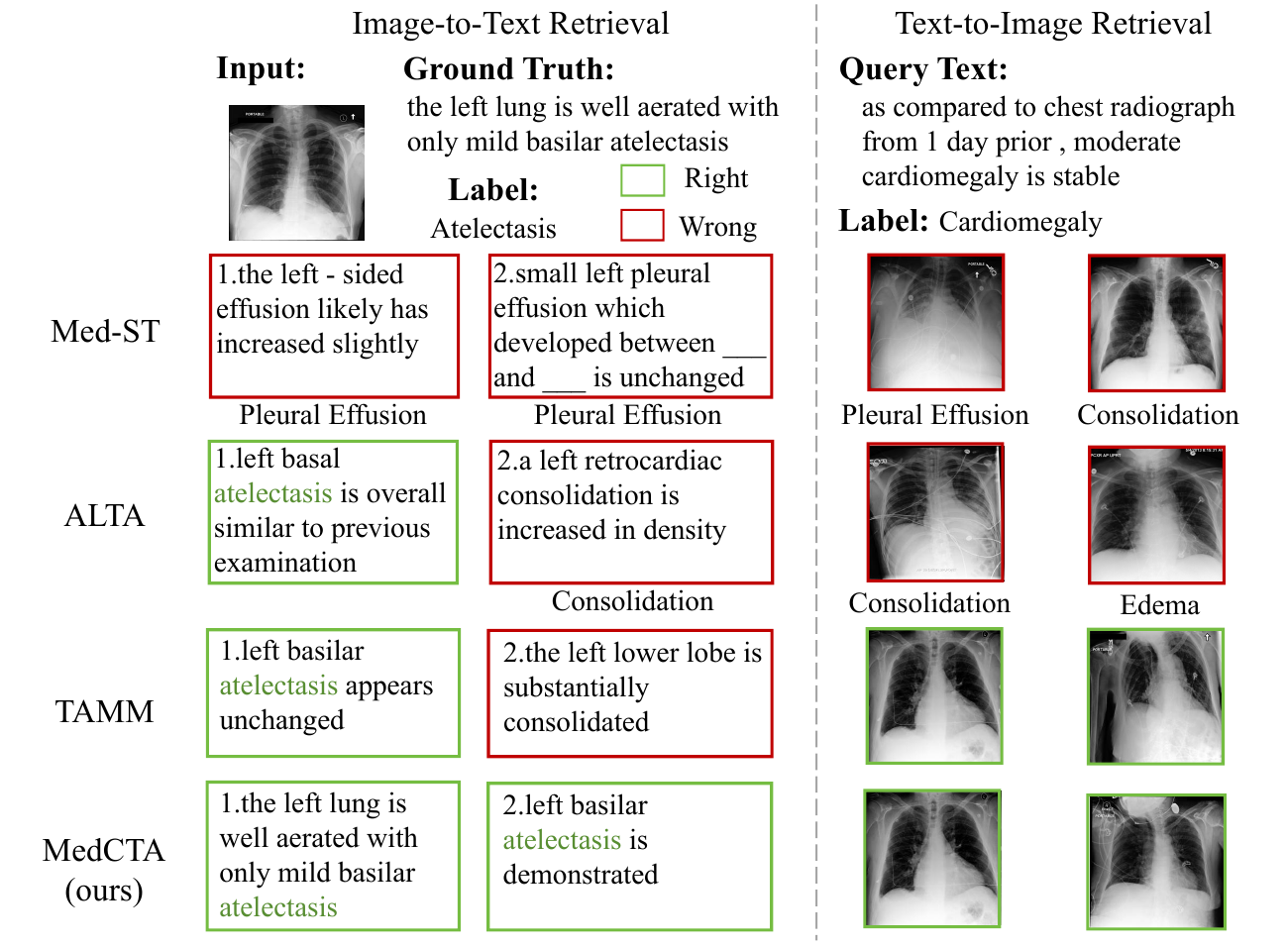}
\caption{
Qualitative examples of image-to-text and text-to-image retrieval on MIMIC-5$\times$200.
For each method, we show the top two retrieved samples given an input chest X-ray or a query sentence.
Green and red boxes denote correct and incorrect retrievals, respectively, with abnormality categories annotated below the incorrect samples.
}
\label{fig:retrieval_image}
\end{figure*}

\textbf{Cross-modal Retrieval on MIMIC-5$\times$200.}
We evaluate text-to-image and image-to-text retrieval on MIMIC-5$\times$200.
As shown in Table~\ref{tab:retrieval_mimic}, MedCTA achieves the best results on seven of eight metrics and ranks second on text-to-image P@2.
Fig.~\ref{fig:retrieval_image} shows qualitative image-to-text and text-to-image examples, where green/red boxes denote correct/incorrect retrievals and labels indicate abnormality categories.
These results demonstrate that combining static alignment with abnormality-conditioned trajectories and patient-course progression improves cross-modal representation learning.

\begin{table*}[!htbp]
\centering
\caption{Cross-modal retrieval results on the MIMIC-CXR 5$\times$200 benchmark. Performance is reported as precision (\%) at top-$k$ (P@$k$). Best results are shown in bold, and second-best results are underlined.}
\small
\renewcommand{\arraystretch}{1}
\begin{tabular*}{\textwidth}{@{\extracolsep{\fill}}lcccccccc@{}}
\toprule
& \multicolumn{4}{c}{\textbf{Text $\rightarrow$ Image}}
& \multicolumn{4}{c}{\textbf{Image $\rightarrow$ Text}} \\
\cmidrule(lr){2-5} \cmidrule(lr){6-9}
\textbf{Method} & \textbf{P@1} & \textbf{P@2} & \textbf{P@5} & \textbf{P@10}
& \textbf{P@1} & \textbf{P@2} & \textbf{P@5} & \textbf{P@10} \\
\midrule
MGCA          & 74.50 & 76.30 & 71.16 & 60.85 & 63.62 & 63.88 & 64.11 & 61.95 \\
MedCLIP       & 45.75 & 47.10 & 48.63 & 43.07 & 50.31 & 48.37 & 48.30 & 48.09 \\
PRIOR         & 47.13 & 48.11 & 47.53 & 47.24 & 49.50 & 52.55 & 51.95 & 36.55 \\
MAVL          & 50.00 & 50.00 & 42.91 & 37.27 & 50.00 & 50.00 & 49.47 & 50.00 \\
CARZero       & 52.44 & 50.00 & 49.47 & 50.01 & 50.00 & 50.00 & 51.65 & 47.38 \\
Med-ST        & 75.25 & 76.18 & 71.36 & \underline{65.04} & 64.50 & 63.97 & 62.48 & 59.00 \\
ALTA          & 69.80 & 65.85 & 52.60 & 37.72 & 56.80 & 55.40 & 46.80 & 46.80 \\
TAMM & \underline{78.69} & \textbf{79.26} & \underline{73.07} & 64.03 & \underline{66.44} & \underline{67.63} & \underline{66.05} & \underline{62.50} \\
\textbf{Ours} & \textbf{80.24} & \underline{79.08} & \textbf{75.68} & \textbf{65.82} & \textbf{69.13} & \textbf{68.63} & \textbf{67.23} & \textbf{63.01} \\
\bottomrule
\end{tabular*}
\label{tab:retrieval_mimic}
\end{table*}

\textbf{Zero-shot Classification Tasks.}
We evaluate the transferability of MedCTA on RSNA Pneumonia and COVIDx under a zero-shot setting.
As shown in Table~\ref{tab:zero_shot}, MedCTA achieves the best results on both datasets, with 86.05\% accuracy and 79.25\% F1 on RSNA Pneumonia, and 95.66\% accuracy and 88.48\% F1 on COVIDx.
This indicates that modeling longitudinal clinical change enhances visual representation learning.
By integrating static image--report alignment with abnormality-conditioned trajectories and patient-course progression, MedCTA learns more transferable visual features for zero-shot disease classification.

\begin{table*}[!t]
\centering
\caption{Zero-shot classification results on the RSNA Pneumonia and COVIDx datasets. Performance is reported in terms of Accuracy (ACC, \%) and F1 (\%). Best results are shown in bold, and second-best results are underlined.}
\small
\setlength{\tabcolsep}{4pt}
\renewcommand{\arraystretch}{1}
\begin{tabular*}{\textwidth}{@{\extracolsep{\fill}}lcccc@{}}
% \begin{tabular*}{0.85\textwidth}{@{\extracolsep{\fill}}lcccc@{}}
\toprule
& \multicolumn{2}{c}{\textbf{RSNA Pneumonia}} 
& \multicolumn{2}{c}{\textbf{COVIDx}} \\
\cmidrule(lr){2-3} \cmidrule(lr){4-5}
\textbf{Method} 
& \textbf{ACC} & \textbf{F1} 
& \textbf{ACC} & \textbf{F1} \\
\midrule
MGCA          & 85.06 $\pm$ 0.05 & 76.56 $\pm$ 0.10 & 92.13 $\pm$ 0.10 & 80.10 $\pm$ 0.65 \\
MedCLIP       & 84.17 $\pm$ 0.16 & 77.38 $\pm$ 0.35 & 90.19 $\pm$ 0.09 & 75.60 $\pm$ 2.10 \\
PRIOR         & 81.30 $\pm$ 0.35 & 72.02 $\pm$ 1.09 & 90.01 $\pm$ 0.19 & 87.97 $\pm$ 0.68 \\
MAVL          & 79.99 $\pm$ 0.83 & 73.14 $\pm$ 0.48 & 88.77 $\pm$ 0.92 & 85.09 $\pm$ 1.71 \\
CARZero       & 84.28 $\pm$ 0.07 & 76.94 $\pm$ 0.40 & 91.92 $\pm$ 0.20 & 85.83 $\pm$ 0.98 \\
Med-ST        & 85.15 $\pm$ 0.04 & 76.96 $\pm$ 0.25 & 92.27 $\pm$ 0.17 & 80.22 $\pm$ 0.35 \\
ALTA          & 84.24 $\pm$ 0.20 & 76.54 $\pm$ 0.61 & 91.53 $\pm$ 0.19 & 77.62 $\pm$ 1.26 \\
TAMM          & \underline{85.48 $\pm$ 0.05} & \underline{78.21 $\pm$ 0.14} & \underline{93.03 $\pm$ 0.06} & \underline{87.32 $\pm$ 0.38} \\
\textbf{Ours} & \textbf{86.05 $\pm$ 0.03} & \textbf{79.25 $\pm$ 0.08} & \textbf{95.66 $\pm$ 0.12} & \textbf{88.48 $\pm$ 0.16} \\
\bottomrule
\end{tabular*}
\label{tab:zero_shot}
\end{table*}

\subsection{Ablation Study}

\textbf{Contribution of Different Components.}
We ablate $\mathcal{L}_{\mathrm{clip}}$, $\mathcal{L}_{\mathrm{abn}}$, and $\mathcal{L}_{\mathrm{pat}}$ to assess the contribution of each objective.
As shown in Table~\ref{tab:ablation_mscxrt_temporal}, removing any objective consistently reduces performance on temporal sentence similarity and temporal image classification, indicating their complementary effects.
% The full model achieves the best results, demonstrating the benefit of combining static alignment with dual-scope clinical change supervision.
The full model performs best, demonstrating the effectiveness of combining static alignment with dual-scope clinical change supervision.
More detailed ablations are reported in Appendix~\ref{Appendix_Different_Components}.

\begin{table*}[!t]
\centering
\caption{
Ablation study on temporal tasks on the MS-CXR-T dataset.  
Accuracy (\%) for temporal sentence similarity and temporal image classification. 
Relative changes from the full model are shown below.
}
\scriptsize 
\setlength{\tabcolsep}{4pt}
\renewcommand{\arraystretch}{1}
\begin{tabular}{lcccccc}
\toprule
\textbf{Method} 
& \textbf{Temporal Sentence Similarity} 
& \multicolumn{5}{c}{\textbf{Temporal Image Classification}} \\
& 
& Consolidation & Edema & Pleural Effusion & Pneumonia & Pneumothorax \\
\midrule
\textbf{Ours} 
& \textbf{95.72} 
& \textbf{67.14} & \textbf{70.30} & \textbf{65.59} & \textbf{72.85} & \textbf{61.66} \\
\midrule
\textbf{w/o clip loss} 
& 92.24 
& 59.58 & 59.51 & 63.44 & 67.50 & 56.75 \\
& {($-$3.48)} 
& {($-$7.56)} & {($-$10.79)} & {($-$2.15)} & {($-$5.35)} & {($-$4.91)} \\
\textbf{w/o abnormality loss} 
& 87.47 
& 55.22 & 51.46 & 58.41 & 65.24 & 53.33 \\
& {($-$8.25)} 
& {($-$11.92)} & {($-$18.84)} & {($-$7.18)} & {($-$7.61)} & {($-$8.33)} \\
\textbf{w/o patient loss} 
& 90.55 
& 56.98 & 57.66 & 59.58 & 65.80 & 54.18 \\
& {($-$5.17)} 
& {($-$10.16)} & {($-$12.64)} & {($-$6.01)} & {($-$7.05)} & {($-$7.48)} \\
\bottomrule
\end{tabular}

\label{tab:ablation_mscxrt_temporal}
\end{table*}

\textbf{Effect of Query Initialization.}
We compare random and clinically initialized abnormality queries in Appendix~\ref{Appendix_Query}.
Clinical initialization consistently performs better, indicating that medical prompts provide useful semantic priors for constructing abnormality-specific trajectories.

\textbf{Effect of Different LLM Parsers.}
MedCTA uses Qwen2-7B~\cite{bai2023qwen} as the offline parser for abnormality-specific and patient-course trend supervision. To assess parser robustness, we replace it with Llama 3-8B~\cite{grattafiori2024llama}.
Appendix~\ref{Appendix_LLM} shows stable image-text retrieval performance, suggesting that MedCTA's gains come from structured temporal supervision that transforms implicit longitudinal semantics into learnable clinical change signals, rather than from a specific LLM backbone.

\section{Conclusion}

We introduced MedCTA, a clinical trajectory alignment framework that extends medical vision--language pre-training from visit-level image--report matching to longitudinal clinical change learning. MedCTA models change at two complementary scopes: abnormality-specific trajectories capture heterogeneous abnormality evolution, while patient-course progression captures overall clinical status. With structured trend supervision extracted from longitudinal reports by an offline LLM parser, MedCTA enables trajectory-level learning without manual temporal annotations while preserving static image--report alignment.
Experiments on temporal image classification, image--text retrieval, and zero-shot classification show consistent gains over strong medical VLP baselines, demonstrating the value of modeling clinical change through both abnormality-specific evolution and patient-course progression.

% \section*{References}

{
\small
\bibliographystyle{unsrt}
\bibliography{references}
}

%%%%%%%%%%%%%%%%%%%%%%%%%%%%%%%%%%%%%%%%%%%%%%%%%%%%%%%%%%%%

\appendix

\section{Additional Experimental Details}
\label{Appendix_A}

\subsection{Pretraining Data}
We pretrain MedCTA on MIMIC-CXR~\cite{johnson2019mimic}, which contains chest X-ray images and corresponding radiology reports. For longitudinal modeling, we reorganize the data into patient-level sequences ordered by study time. Following prior temporal medical VLP settings~\cite{DBLP:conf/icml/Yang00W24}, each sequence includes up to four consecutive image--report pairs, while shorter sequences are preserved. The dataset is publicly available at \url{https://physionet.org/content/mimic-cxr-jpg/2.1.0/}.

\subsection{Downstream Tasks}
We evaluate MedCTA on downstream benchmarks disjoint from the pre-training data, covering both longitudinal temporal reasoning and cross-modal understanding.

\textbf{Longitudinal temporal reasoning:} We evaluate three tasks. \textit{Temporal image classification} is fine-tuned on Chest ImaGenome~\cite{wu2021chest} and tested on MS-CXR-T~\cite{bannur2023learning}, reporting macro-averaged accuracy over progression types. \textit{Temporal sentence similarity classification} is evaluated as a zero-shot binary classification on MS-CXR-T sentence pairs, reporting accuracy.
\textit{Dynamic phrase grounding} is assessed through qualitative visualization on Chest ImaGenome, where we examine whether the model localizes image regions relevant to temporal progression descriptions across consecutive chest X-rays.

\textbf{Cross-modal understanding:} We evaluate \textit{cross-modal retrieval} on MIMIC-5$\times$200~\cite{MedCLIP}, reporting P@1, P@2, P@5, and P@10 for both image-to-text and text-to-image retrieval. We also assess \textit{zero-shot classification} on COVIDx~\cite{wang2020covid} and RSNA Pneumonia~\cite{shih2019augmenting} using Accuracy and F1.

\subsection{Baselines}
We compare MedCTA with recent representative medical vision--language methods, grouped according to whether they explicitly model temporal information. 
\textit{Non-temporal vision--language baselines} focus on static image--text alignment: MGCA~\cite{wang2022multi} aligns radiographs and reports at multiple granularities; MedCLIP~\cite{MedCLIP} improves contrastive learning with semantic-aware matching; CARZero~\cite{CARZero} leverages LLM-based prompts and cross-attention for zero-shot classification; PRIOR~\cite{cheng2023prior} combines cross-modal reconstruction with global-local supervision; and MAVL~\cite{phan2024decomposing} uses LLM-guided disease decomposition to enhance attribute-level grounding. 
In contrast, \textit{temporal baselines} explicitly incorporate longitudinal context: Med-ST~\cite{DBLP:conf/icml/Yang00W24} introduces temporal supervision through historical image--text pairs and cycle consistency, while ALTA~\cite{lian2025efficient} models temporal views with masked pretraining and efficient adaptation. TAMM~\cite{bai2026medical} leverages LLM-derived trend labels and rationales from adjacent reports to inject temporal supervision into medical vision-language pretraining.

\subsection{Implementation Details}
\label{subsec:implementation_details}
We initialize MedCTA with pretrained MGCA weights to improve optimization stability and accelerate convergence. The model is trained for 8 epochs on 4 NVIDIA A40 GPUs. We use the AdamW optimizer with a weight decay of $1 \times 10^{-6}$ and an initial learning rate of $1 \times 10^{-5}$. A cosine annealing schedule with 40\% warm-up is adopted, gradually decaying the learning rate to $1 \times 10^{-8}$.
All images are resized to $256 \times 256$, followed by a random crop to $224 \times 224$ for data augmentation. 
For downstream evaluation, we either use the pretrained model in a zero-shot manner or fine-tune a lightweight task-specific head, depending on the benchmark. 
For cross-modal retrieval, temporal sentence similarity classification, and dynamic phrase grounding, we directly evaluate the pretrained model without additional training, and thus do not report standard deviations.
In contrast, zero-shot classification and temporal image classification involve training a lightweight classifier on top of pretrained representations. For these tasks, we run experiments with three different random seeds and report the mean and standard deviation to reflect training variability. This evaluation protocol follows prior work for fair comparison.

\section{Ablation experimental results}
\label{appendix_Ablation_experimental_results}

In this section, we provide additional ablation results to further analyze the contribution of each component in MedCTA, the effect of abnormality query initialization, and the robustness to different LLM-based trend supervision extractors.

\subsection{Contribution of Different Components.}
\label{Appendix_Different_Components}
To further validate the effectiveness of the three training objectives, we report additional ablation results on image--text retrieval. Specifically, we remove the static image--report alignment loss $\mathcal{L}_{\mathrm{clip}}$, the abnormality-specific trend loss $\mathcal{L}_{\mathrm{abn}}$, and the patient-course trend loss $\mathcal{L}_{\mathrm{pat}}$ from the full objective, respectively.
As shown in Table~\ref{tab:ablation_retrieval_appendix}, removing each objective generally degrades retrieval performance across most metrics, indicating that the three objectives provide complementary supervision. Removing $\mathcal{L}_{\mathrm{clip}}$ causes the largest drop, suggesting that visit-level image--report alignment remains essential for cross-modal retrieval. Removing $\mathcal{L}_{\mathrm{abn}}$ also reduces performance, reflecting the benefit of modeling fine-grained abnormality evolution. Removing $\mathcal{L}_{\mathrm{pat}}$ leads to smaller but mostly consistent declines, suggesting that patient-course progression modeling further improves the learned representations. These retrieval results are broadly consistent with the temporal reasoning results in the main paper and support the complementary roles of the three objectives.

\begin{table*}[t]
\centering
\caption{
Additional ablation results on image--text retrieval on the MIMIC-CXR 5$\times$200 benchmark. 
Performance is reported as precision (\%) at top-$k$ (P@$k$). 
Relative changes from the full model are shown below.
}
\scriptsize
\setlength{\tabcolsep}{4pt}
\renewcommand{\arraystretch}{1.1}
\begin{tabular*}{\textwidth}{@{\extracolsep{\fill}}lcccccccc@{}}
\toprule
\textbf{Method} 
& \multicolumn{4}{c}{\textbf{Text $\rightarrow$ Image}} 
& \multicolumn{4}{c}{\textbf{Image $\rightarrow$ Text}} \\
\cmidrule(lr){2-5} \cmidrule(lr){6-9}
& \textbf{P@1} & \textbf{P@2} & \textbf{P@5} & \textbf{P@10} 
& \textbf{P@1} & \textbf{P@2} & \textbf{P@5} & \textbf{P@10} \\
\midrule
\textbf{MedCTA} 
& \textbf{80.24} & \textbf{79.08} & \textbf{75.68} & \textbf{65.82} 
& \textbf{69.13} & \textbf{68.63} & \textbf{67.23} & \textbf{63.01} \\
\midrule

\textbf{w/o $\mathcal{L}_{\mathrm{clip}}$} 
& 58.05 & 55.35 & 54.32 & 46.06
& 53.22 & 53.19 & 53.98 & 48.72 \\
& {($-$22.19)} & {($-$23.73)} & {($-$21.36)} & {($-$19.76)}
& {($-$15.91)} & {($-$15.44)} & {($-$13.25)} & {($-$14.29)} \\

\textbf{w/o $\mathcal{L}_{\mathrm{abn}}$} 
& 77.67 & 76.59 & 73.45 & 63.93
& 67.51 & 66.36 & 64.75 & 61.28 \\
& {($-$2.57)} & {($-$2.49)} & {($-$2.23)} & {($-$1.89)}
& {($-$1.62)} & {($-$2.27)} & {($-$2.48)} & {($-$1.73)} \\

\textbf{w/o $\mathcal{L}_{\mathrm{pat}}$} 
& 78.90 & 78.02 & 74.01 & 65.00
& 68.21 & 68.67 & 66.51 & 61.80 \\
& {($-$1.34)} & {($-$1.06)} & {($-$1.67)} & {($-$0.82)}
& {($-$0.92)} & {($+$0.04)} & {($-$0.72)} & {($-$1.21)} \\
\bottomrule
\end{tabular*}
\label{tab:ablation_retrieval_appendix}
\end{table*}

\subsection{Effect of Query Initialization.}
\label{Appendix_Query}
We further investigate whether the abnormality queries benefit from clinical semantic initialization. We compare two strategies: random initialization and clinical semantic initialization using abnormality-specific medical prompts. 
As shown in Table~\ref{query_init}, clinical semantic initialization improves most retrieval metrics compared with random initialization.
This suggests that abnormality prompts provide useful semantic priors, helping the queries select abnormality-relevant visual and textual evidence and construct more effective abnormality-specific progression trajectories.

\begin{table*}[t]
\centering
\caption{Effect of abnormality query initialization on image--text retrieval on the MIMIC-CXR 5$\times$200 benchmark. Performance is reported as precision at top-$k$ (P@k).}
\scriptsize
\setlength{\tabcolsep}{4pt}
\renewcommand{\arraystretch}{1.1}
\begin{tabular*}{\textwidth}{@{\extracolsep{\fill}}lcccccccc@{}}
\toprule
\textbf{Method} 
& \multicolumn{4}{c}{\textbf{Text $\rightarrow$ Image}} 
& \multicolumn{4}{c}{\textbf{Image $\rightarrow$ Text}} \\
\cmidrule(lr){2-5} \cmidrule(lr){6-9}
& \textbf{P@1} & \textbf{P@2} & \textbf{P@5} & \textbf{P@10} 
& \textbf{P@1} & \textbf{P@2} & \textbf{P@5} & \textbf{P@10} \\
\midrule

\textbf{Random init} 
&78.37   &77.69   &72.94   &65.32   
&68.27   &65.42   &68.15   &61.82   \\

\textbf{Clinical init} 
& 80.24 & 79.08 & 75.68 & 65.82
& 69.13 & 68.63 & 67.23 & 63.01 \\

\bottomrule
\end{tabular*}
\label{query_init}
\end{table*}

\subsection{Effect of Different LLM Parsers.}
\label{Appendix_LLM}
MedCTA uses an offline LLM parser to extract structured trend supervision from longitudinal reports. To examine whether the performance depends on a specific LLM backbone, we replace Qwen2-7B~\cite{bai2023qwen} with Llama 3-8B~\cite{grattafiori2024llama} while keeping the remaining training and evaluation settings unchanged.
As shown in Table~\ref{differ_LLM}, MedCTA maintains stable performance under different LLM settings on image--report retrieval. These results indicate that MedCTA does not rely on a particular LLM backbone. Instead, its gains mainly come from the structured temporal supervision that converts implicit longitudinal report semantics into learnable clinical change signals.

\begin{table*}[!htbp]
\centering
\caption{Robustness to different LLM parsers on image--text retrieval on the MIMIC-CXR 5$\times$200 benchmark. Performance is reported as precision at top-$k$ (P@k).}
\label{differ_LLM}
\vspace{0.5em}
\scriptsize
\setlength{\tabcolsep}{4pt}
\renewcommand{\arraystretch}{1.1}
\begin{tabular*}{\textwidth}{@{\extracolsep{\fill}}lcccccccc@{}}
\toprule
\textbf{Method} 
& \multicolumn{4}{c}{\textbf{Text $\rightarrow$ Image}} 
& \multicolumn{4}{c}{\textbf{Image $\rightarrow$ Text}} \\
\cmidrule(lr){2-5} \cmidrule(lr){6-9}
& \textbf{P@1} & \textbf{P@2} & \textbf{P@5} & \textbf{P@10} 
& \textbf{P@1} & \textbf{P@2} & \textbf{P@5} & \textbf{P@10} \\
\midrule
\textbf{Qwen2-7B} 
& 80.24 & 79.08 & 75.68 & 65.82
& 69.13 & 68.63 & 67.23 & 63.01 \\
\textbf{Llama 3-8B} 
& 80.11 & 78.92 & 76.89 & 66.09 
& 69.56 & 68.54 & 66.85 & 63.14 \\
\bottomrule
\end{tabular*}
\end{table*}

\section{Limitations and Broader Impact}
\label{app:limitations_impact_assets}

\paragraph{Limitations.}
Although MedCTA shows consistent improvements across temporal reasoning and cross-modal understanding tasks, it still has several limitations. 
First, MedCTA relies on structured trend supervision extracted by an offline LLM parser from longitudinal radiology reports. 
Although our experiments with different LLM parsers show stable performance, the quality of the extracted trend labels may still be affected by the reasoning ability, prompting strategy, and possible parsing errors of the LLM. 
Second, MedCTA models abnormality-specific trajectories based on a predefined abnormality set, which may not fully cover rare diseases, fine-grained clinical findings, or complex comorbid conditions. 
Third, our experiments are mainly conducted on chest X-ray datasets and radiology reports. 
Further validation on other imaging modalities, medical domains, institutions, and real-world clinical workflows remains an important direction for future work. 
Finally, MedCTA is designed for representation learning and should not be directly used for clinical diagnosis or decision-making without careful validation and expert supervision.

\paragraph{Broader Impact.}
MedCTA aims to improve medical vision--language representation learning by explicitly modeling longitudinal clinical change. 
Its potential positive impacts include enhancing temporally aware medical image understanding, improving progression-related retrieval and grounding, and supporting research on longitudinal disease modeling. 
By reducing the need for manual temporal annotations through offline LLM-derived supervision, MedCTA may also facilitate the scalable use of longitudinal medical records for representation learning. 
However, the method may inherit biases and reporting inconsistencies from the underlying medical datasets and radiology reports. 
In addition, inaccurate LLM-derived trend labels could affect the learned representations, especially in underrepresented diseases or patient groups. 
Therefore, deployment in high-stakes clinical scenarios should carefully consider data privacy, fairness, and robustness.

%%%%%%%%%%%%%%%%%%%%%%%%%%%%%%%%%%%%%%%%%%%%%%%%%%%%%%%%%%%%

\newpage

\end{document}